\documentclass[twoside]{article}

\PassOptionsToPackage{table,dvipsnames}{xcolor}
\usepackage{amazon_paper}

\usepackage{amssymb}
\usepackage{booktabs}
\usepackage[]{graphicx}
\usepackage{caption}
\usepackage{subcaption}
\usepackage{wrapfig}

\usepackage{multirow}
\usepackage{array}
\definecolor{baselinegray}{gray}{0.92}
\definecolor{overallbg}{RGB}{238,242,255}
\newcolumntype{O}{>{\columncolor{baselinegray}}c}

\usepackage{amsmath}
\usepackage{enumitem}
\usepackage{algorithm}
\usepackage{algpseudocode}

\usepackage{pifont}
\usepackage{makecell} 

\usepackage{listings}

\lstdefinelanguage{json}{
    basicstyle=\ttfamily\small,
    comment=[l]{//},
    morestring=[b]",
    stringstyle=\color{blue},
    literate=
     *{0}{{{\color{black}0}}}{1}
      {1}{{{\color{black}1}}}{1}
      {2}{{{\color{black}2}}}{1}
      {3}{{{\color{black}3}}}{1}
      {4}{{{\color{black}4}}}{1}
      {5}{{{\color{black}5}}}{1}
      {6}{{{\color{black}6}}}{1}
      {7}{{{\color{black}7}}}{1}
      {8}{{{\color{black}8}}}{1}
      {9}{{{\color{black}9}}}{1}
}

\usepackage[most]{tcolorbox}
\usepackage{courier} 

\newtcolorbox{promptbox}[2][]{%
  enhanced,
  breakable,
  colback=gray!2,
  colframe=gray!30,
  boxrule=0.4pt,
  arc=6pt,
  outer arc=2pt,
  fontupper=\ttfamily\small,
  title={#2},
  coltitle=white,
  colbacktitle=gray!80!black,
  attach boxed title to top left={yshift=-1.5mm, xshift=2mm},
  boxed title style={
    sharp corners,
    boxrule=0pt,
    colback=gray!80!black,
    colupper=white,
    fontupper=\bfseries\footnotesize,
  },
  #1
}

\begin{document}

\title{Reinforcing the Generation Order of Multimodal\\ Masked Diffusion Models}

\author{
\small{Yidong Ouyang$^{1}$\thanks{Work done during an internship at Amazon.}, Zhe Wang$^{2}$, Sourav~Bhabesh$^{2}$, Dmitriy~Bespalov$^{2}$}
  \\[1em]
  {\fontsize{10pt}{11pt}\selectfont
$^{1}$University of California, Los Angeles \quad
$^{2}$AGI Foundations for AWS}
}

\maketitle

\begin{abstract}
Diffusion Language Models (DLMs) have recently achieved substantial progress in natural language generation tasks. Recent research demonstrates that adaptive token generation ordering can significantly improve performance in mathematical reasoning and code synthesis applications. In this work, we investigate the optimization of generation order for both text-to-image synthesis and multimodal understanding. We first establish that, unlike structured problems in language generation such as Sudoku puzzles, model logits alone are insufficient for determining optimal generation sequences in text-to-image generation and multimodal understanding. To address this challenge, we introduce a learnable control module trained via Group Relative Policy Optimization (GRPO) to determine the generation order. Our results demonstrate that learning this control block substantially improves both text-to-image alignment and multimodal understanding in DLMs. In particular, it enhances the model’s ability to capture fine-grained spatial relationships in generated images while also strengthening performance on multimodal reasoning and comprehension tasks. We evaluate our framework on GenEval, an object-focused benchmark for text-to-image alignment, where it achieves 4.08\% relative improvements. In addition, experiments on VLMEvalKit confirm 4.85\% relative improvements in multimodal understanding, highlighting the broad effectiveness of our approach.

\end{abstract}

\section{Introduction}
\label{sec:intro}
Masked diffusion models (MDMs) \citep{Austin2021StructuredDD, Hoogeboom2021ArgmaxFA, Sun2022ScorebasedCD, Lou2023DiscreteDM, Arriola2025BlockDI, Ou2024YourAD, Rutte2025GeneralizedID, Shi2024SimplifiedAG, Nie2025LargeLD, Gat2024DiscreteFM, Campbell2024GenerativeFO} have recently emerged as a powerful class of generative models for discrete data, achieving promising success in language generation tasks. Unlike traditional autoregressive models, which follow a fixed left-to-right decoding scheme, MDMs are trained to solve any-order infilling problems and can generate tokens in essentially arbitrary orders during inference. This flexibility allows MDMs to better handle partially observed inputs and to exploit non-sequential dependencies among tokens.

As MDMs gained popularity, the community has begun to recognize that the generation order—the sequence in which unobserved tokens are produced—plays a crucial role in determining the final output quality \citep{Kim2025TrainFT, Huang2025ReinforcingTD}. Early approaches typically selected generation positions at random, but recent work has shown that order selection can be made more informed by leveraging intermediate model predictions. For example, Zheng et al. \citep{Zheng2023ARD} introduced the Top-$K$ strategy, which prioritizes positions with the highest predicted token probabilities. Similarly, \citet{Kim2025TrainFT} proposed the Top-$K$ margin strategy, ranking positions by the absolute difference between the top two predicted probabilities. These strategies, by exploiting the model’s confidence estimates, significantly improve performance in structured reasoning tasks such as Sudoku solving.

However, existing studies on generation order have been confined to language generation tasks. With the recent development of multimodal masked diffusion models \citep{Yang2025MMaDAML}, which extend MDMs to text-to-image generation and multimodal understanding, a natural question arises: Does the generation order matter in multimodal settings, and if so, how should it be determined? Our preliminary experiments reveal that strategies such as the Top-$K$ margin, while highly effective for symbolic reasoning, fail to yield improvements in image generation quality and multimodal understanding. This discrepancy highlights an important gap: multimodal tasks require order control mechanisms that go beyond simple confidence-based heuristics.

To address this, we propose a control block for dynamically determining the generation order in multimodal masked diffusion models. Unlike heuristic approaches that rely solely on logits, the control block is learnable and can be trained to optimize the generation process directly via Group Relative Policy Optimization (GRPO). This enables the model to adapt its generation order policy to the specific requirements of image synthesis, capturing complex interdependencies between spatially distributed visual tokens.

We evaluate our approach on the GenEval benchmark, a challenging suite designed for rigorous assessment of text-to-image generation, and VLMEvalKit benchmark, a comprehensive evaluation framework for multimodal understanding covering visual question answering, reasoning, and perception. Experimental results demonstrate that our method outperforms existing order selection strategies in this setting, confirming the importance of learning-based control for multimodal generation.

Our contributions are as follows:

\begin{itemize}[leftmargin=1.5em]
\item We observe existing confidence-based strategies do not yield improvement in image generation quality and multimodal reasoning.

\item We propose a control block to determine the generation order and optimize it via Group Relative Policy Optimization (GRPO).

\item We provide extensive empirical evaluation on the GenEval benchmark and VLMEvalKit benchmark with 4.08\%  and 4.85\% relative improvement respectively, demonstrating the effectiveness and generality of our approach.
\end{itemize}

\section{Preliminary}
\label{sec:preli}
In this section, we first introduce the background of masked diffusion models and then summarize a widely used  Group Relative Policy Optimization.

\subsection{Masked Diffusion Model}

Masked diffusion models (MDMs) \citep{Austin2021StructuredDD, Hoogeboom2021ArgmaxFA, Sun2022ScorebasedCD, Lou2023DiscreteDM, Arriola2025BlockDI, Ou2024YourAD, Rutte2025GeneralizedID, Shi2024SimplifiedAG, Nie2025LargeLD} are defined by a forward (noising) process and a reverse (denoising) process. In the forward process, tokens in a sequence are progressively replaced with a special mask token until the sequence is fully masked. The reverse process learns to reconstruct the original sequence starting from the fully masked state.

\paragraph{Forward process}
Given a clean sequence $x_T \in \mathcal{S}^{\mathcal{L}}$ sampled from the data distribution
$p_{\mathrm{data}}$\footnote{We use $x_T$ to denote the clean data sample instead of the more common convention where $x_0$ denotes the clean data sample. This notation is adopted to align with the GRPO-based optimization.}, where $\mathcal{S}$ denotes the finite token space, and a noise level
$t\in\{0,1, \cdots, T\}$, the forward process is defined as
$x_t\sim q_{t|T}(\cdot|x_T)$.
The corruption process independently masks each coordinate:
\[
q_{t|T}(x_t|x_T)
=
\prod_{i=1}^{\mathcal{L}}
q_{t|T}(x_t^i|x_T^i),
\]
where
\[
q_{t|T}(x_t^i|x_T^i)
=
\operatorname{Cat}
\left(
\alpha_t e_{x_T^i}
+
(1-\alpha_t)e_m
\right).
\]
Here, $\alpha_t$ is a predefined noise schedule satisfying
$\alpha_0=0$ and $\alpha_T=1$.
The vector $e_{x_T^i}$ denotes the one-hot representation of token $x_T^i$,
and $e_m$ denotes the one-hot representation of the mask token $m$.
Equivalently, each token remains unchanged with probability $\alpha_t$ and
is replaced by the mask token with probability $1-\alpha_t$.

\paragraph{Reverse process}
The reverse transition conditioned on the original
sequence is defined as
$q_{t|t-1}(x_t|x_{t-1},x_T)$.
Due to the independence across coordinates, it factorizes as
\[
q_{t|t-1}(x_t|x_{t-1},x_T)
=
\prod_{i=1}^{\mathcal{L}}
q_{t|t-1}(x_t^i|x_{t-1}^i,x_T^i),
\]
where
\[
q_{t|t-1}(x_t^i|x_{t-1}^i,x_T^i)
=
\begin{cases}
\operatorname{Cat}(e_{x_{t-1}^i}),
&
x_{t-1}^i\neq m,
\\[6pt]
\operatorname{Cat}
\left(
\frac{1-\alpha_t}{1-\alpha_{t-1}}e_m
+
\frac{\alpha_t-\alpha_{t-1}}{1-\alpha_{t-1}}e_{x_T^i}
\right),
&
x_{t-1}^i=m .
\end{cases}
\]

\paragraph{Training objective}
The model is trained by minimizing the weighted cross-entropy objective
\[
\mathcal{L}_{\theta}
=
-\sum_{t=1}^T
\frac{\alpha_t-\alpha_{t-1}}{1-\alpha_t}
\,
\mathbb{E}_{\substack{
x_T\sim p_{\mathrm{data}}\\
x_t\sim q_{t|T}(\cdot|x_T)
}}
\left[
\sum_{i=1}^{\mathcal{L}}
\delta_m(x_t^i)
e_{x_T^i}^\top
\log p_\theta(x_T^i|x_t,t)
\right],
\]
where $\delta_x(z)$ denotes Kronecker delta satisfying $\delta_x(z)=1$ if $x=z$. In practice, the denoising model is often implemented without explicit time conditioning, i.e., $p_\theta(x_T|x_t, t) = p_\theta(x_T|x_t)$, since the masked input $x_t$ implicitly encodes the noise level $t$ through the proportion of masked tokens.

\paragraph{Sampling process}
Sampling proceeds by simulating the reverse process from $t=0$ to $t=T$. Starting from a fully masked sequence $x_0$, we iteratively generate $x_{t_{k}}$ from $x_{t_{k-1}}$.
At each step, the model first predicts the clean sequence
\[
\hat{x}_T \sim p_\theta(\cdot \mid x_{t_{k-1}}),
\]
and then samples
\[
x_{t_{k}} \sim q_{t_{k} \mid t_{k-1}}\big(\cdot \mid x_{t_{k-1}}, \hat{x}_T \big).
\]

The transition is applied independently across coordinates, where only masked positions are updated. Repeating this procedure progressively removes masks and yields sample $x_T$.

\subsection{GRPO}

Policy gradient methods have been widely adopted in the post-training stage to enhance the performance of large models \citep{Ouyang2022TrainingLM}. While Proximal Policy Optimization (PPO) \citep{Schulman2017ProximalPO} remains a dominant approach in online RL, it requires learning an additional value function $V$, leading to increased computational overhead. Group Relative Policy Optimization (GRPO) \citep{Shao2024DeepSeekMathPT} provides a more efficient alternative by estimating advantages via group statistics.

In the multimodal setting, we consider a conditioning input $c = (x^{\text{text}}, x^{\text{img}})$, where $x^{\text{text}}$ denotes the textual prompt and $x^{\text{img}}$ denotes the visual input (which may be empty for text-to-image generation). For each conditioning input $c$, GRPO samples a group of $G$ outputs $\{o_1, \ldots, o_G\}$ from the old policy $\pi_{\theta_{\text{old}}}(\cdot \mid c)$. Each output $o_i$ receives a scalar reward $r_i$.
The advantage is defined as
\[
A_i= \frac{r_i - \operatorname{mean}\left(\{r_j\}_{j=1}^G\right)}{\operatorname{std}\left(\{r_j\}_{j=1}^G\right)}.
\]
The GRPO objective is given by
\begin{equation}
\begin{aligned}
\mathcal{L}_{\mathrm{GRPO}}(\theta)
&= \mathbb{E}_{\substack{c \sim \mathcal{D} \\ o_1, \ldots, o_G \sim \pi_{\theta_{\text{old}}}(\cdot \mid c)}}
\Bigg[
    \frac{1}{G} \sum_{i=1}^G 
    \frac{1}{|o_i|} \sum_{k=1}^{|o_i|}
    \min \Big(
        \rho_i^k A_i,\,
        \operatorname{clip}(\rho_i^k,\, 1-\varepsilon,\, 1+\varepsilon) A_i
    \Big) \\
&\hspace{100pt}
    - \beta\, D_{\mathrm{KL}}\!\left(
        \pi_\theta(\cdot \mid c, o_i^{<k}) \,\big\|\, \pi_{\mathrm{ref}}(\cdot \mid c, o_i^{<k})
    \right)
\Bigg], \notag
\end{aligned} 
\end{equation}
where $\rho_i^k = \frac{\pi_\theta(o_i^k \mid c, o_i^{<k})}{\pi_{\theta_{\text{old}}}(o_i^k \mid c, o_i^{<k})}$. The clipping parameter $\varepsilon$ controls the update magnitude, while $\beta$ weights the KL regularization toward the reference policy $\pi_{\text{ref}}$.

\paragraph{Diffusion policy optimization}
A key challenge is estimating likelihoods under masked diffusion policies. In \cite{Zhao2025d1SR}, the author adopt a mean-field approximation of sequence likelihood. Let $\phi^{\pi_\theta}(o^k \mid c')$ and $\phi^{\pi_\theta}(o \mid c')$ denote the estimated per-token and sequence probabilities, respectively, where $c'$ is a noised version of the multimodal condition obtained via a masking operator:
$
c' \sim \operatorname{masking}(c).
$
The resulting diffu-GRPO objective is
\[
\begin{aligned}
\mathcal{L}_{\text{diffu-GRPO}}(\theta)
&= \mathbb{E}_{\substack{c \sim \mathcal{D},\, c' \sim \operatorname{masking}(c) \\
o_1, \ldots, o_G \sim \pi_{\theta_{\text{old}}}(\cdot \mid c)}}
\Bigg[
\frac{1}{G} \sum_{i=1}^G \frac{1}{|o_i|} \sum_{k=1}^{|o_i|}
\min \Big(
    \tilde{\rho}_i^k A_i,\,
    \operatorname{clip}(\tilde{\rho}_i^k, 1-\varepsilon, 1+\varepsilon) A_i
\Big)
\\
&\hspace{100pt}
- \beta\, D_{\mathrm{KL}}\!\left(
\phi^{\pi_\theta}(\cdot \mid c') \,\|\, \phi^{\pi_{\text{ref}}}(\cdot \mid c')
\right)
\Bigg],
\end{aligned}
\]
where $
\tilde{\rho}_i^k =
\frac{\phi^{\pi_\theta}(o_i^k \mid c')}
     {\phi^{\pi_{\theta_{\text{old}}}}(o_i^k \mid c')}.$
Recent work such as \citep{Yang2025MMaDAML} improves this framework by introducing structured noising strategies that selectively mask generated tokens while preserving the multimodal condition. In \cite{Huang2025ReinforcingTD, Wang2025RevolutionizingRL}, trajectory information is further leveraged for post-training of diffusion language models.

\section{Method}
\label{sec:method}

Existing order selection strategies, such as Top-$K$ and 
Top-$K$ margin \citep{Zheng2023ARD, Kim2025TrainFT}, rely 
solely on the model's output logits to rank token positions. 
While effective for structured language tasks such as Sudoku 
solving, we empirically find that these heuristics do not 
transfer to multimodal settings: logit confidence at a given 
position does not reliably reflect its importance to the 
overall image layout or visual coherence. This motivates a 
\emph{learned} control mechanism that can capture the complex 
spatial dependencies inherent in visual token generation.

In this section, we introduce a \emph{control block} that adaptively determines the 
generation order at each denoising step. The control block employs 
the Unmask Policy Module (UPM) \citep{Huang2025ReinforcingTD} to 
assign a ranking score $h_{\theta,t}^i$ to each token $i$ at step 
$t$. Based on these scores, a Plackett-Luce 
policy \citep{Ragain2018ChoosingTR, Niu2012ANP} sequentially samples 
a top-$K$ unmask set $\mathcal{U}_t$ without replacement, such that 
higher-scoring tokens are preferentially unmasked first.

\subsection{Control Block for Generation Order}

\begin{figure}
    \centering
    \includegraphics[width=0.5\linewidth]{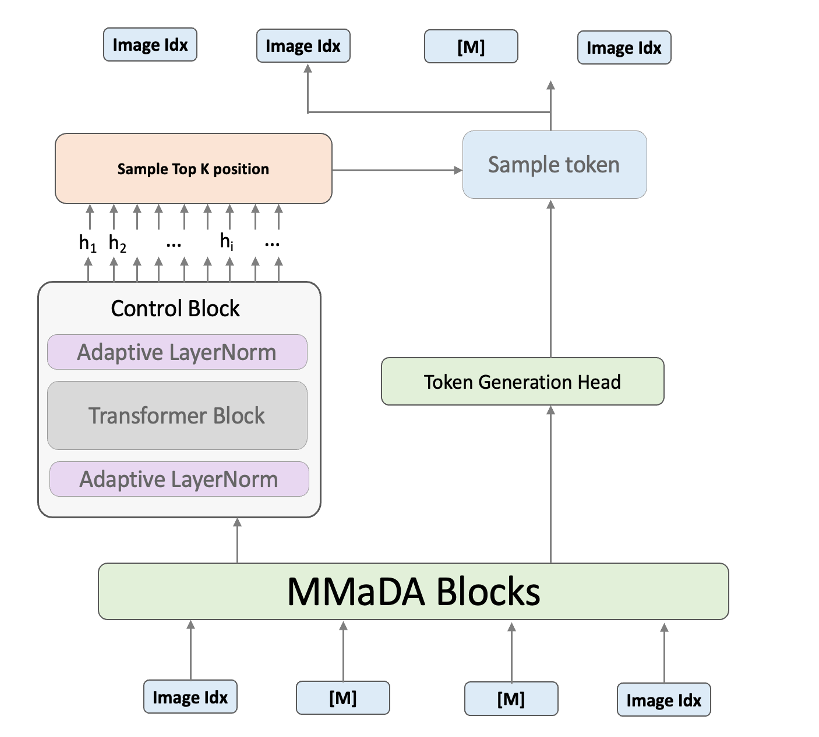}
    \caption{Generation pipeline of the diffusion language model with a control block for determining the generation order.}
    \label{fig:generation_control}
\end{figure}

Let $\mathcal{M}_t$ denote the set of masked tokens remaining after 
step $t$, i.e., $\mathcal{M}_t = \mathcal{M}_{t-1} \setminus 
\mathcal{U}_n$. The probability of selecting a specific unmask set 
$\mathcal{U}_t$ is
\[
\pi_{\theta,t}^{\text{unmask}}(\mathcal{U}_t \mid x_{t-1})
= \prod_{k=1}^K \frac{\exp(h_{\theta,t}^{u_t(k)})}{\sum_{j=k}^K 
\exp(h_{\theta,t}^{u_t(j)}) + \sum_{j \in \mathcal{M}_t} 
\exp(h_{\theta,t}^{u_t(j)})}.
\]
The UPM takes as input the hidden states of the final transformer 
layer at step $t$, together with the step index $t$ and a per-token 
mask indicator, and outputs ranking scores $h_{\theta,t}^i$. The 
$K$ tokens with the highest scores are selected to form the unmask 
set $\mathcal{U}_t$. Token prediction is then performed by the 
main multimodal diffusion model (MMaDA blocks), yielding
\[
\pi_{\theta,t}^{\text{token}}(x_t \mid x_{t-1}, \mathcal{U}_t)
= \prod_{i \in \mathcal{U}_t} p_{\theta,t}(x_t^i \mid x_{t-1}),
\]
where $p_{\theta,t}(x_t^i \mid x_{t-1})$ is a categorical 
distribution over the vocabulary or image patch values. 
The complete step policy factorizes as the product of these 
two components:
\[
\pi_{\theta,t}(x_t \mid x_{t-1})
= \pi_{\theta,t}^{\text{unmask}}(\mathcal{U}_t \mid x_{t-1}) 
\cdot \pi_{\theta,t}^{\text{token}}(x_t \mid x_{t-1}, \mathcal{U}_t).
\]
Figure~\ref{fig:generation_control} illustrates the overall 
generation pipeline, in which the control block governs the 
generation order at each denoising step.

\subsection{Training the Control Block via GRPO}

We formulate the control block as a policy over the reverse 
diffusion process and optimize it via Group Relative Policy 
Optimization (GRPO).

\paragraph{GRPO Objective.}
Given a multimodal input $c$ and $G$ sampled trajectories 
$\{x_t^g\}_{t=0}^T$, each trajectory receives a scalar reward 
$r^g = \texttt{reward\_func}(c, x_T^g)$. Advantages are 
estimated from group statistics:
\[
A^g = \frac{r^g - \mathrm{mean}(r^{1:G})}{\mathrm{std}(r^{1:G})}.
\]
The control policy is optimized by minimizing the 
advantage-weighted likelihood ratio:
\begin{equation}\label{eq:grpo}
\mathcal{L}_{\theta,t}
= -\frac{1}{G} \sum_{g=1}^G
\min\!\Big(
    \rho_{\theta,t}^g\, A^g,\;
    \operatorname{clip}\!\big(\rho_{\theta,t}^g,\, 1-\varepsilon,\, 1+\varepsilon\big)\, A^g
\Big),
\qquad
\rho_{\theta,t}^g = \frac{\pi_{\theta,t}(x_t^g \mid x_{t-1}^g)}{\pi_{\text{old},t}(x_t^g \mid x_{t-1}^g)},
\end{equation}
with gradients accumulated across all diffusion steps:
\[
\nabla_\theta \mathcal{L}_\theta = \sum_{t=1}^T 
\nabla_\theta \mathcal{L}_{\theta,t}.
\]
Unlike the single-step density ratio estimation adopted 
in \cite{Zhao2025d1SR, Yang2025MMaDAML}, we leverage 
full trajectory information by accumulating policy gradients 
over all $N$ denoising steps. This allows the control block 
to receive training signal from the complete generation 
trajectory rather than a single transition, enabling more 
effective optimization of the generation order.

During inference, the control block and the diffusion model operate jointly at each denoising step. At step $t$, the control block first 
computes ranking scores $h_{\theta,t}^i$ for all masked tokens in 
$\mathcal{M}_{t-1}$, and samples a subset $\mathcal{U}_t$ of $K$ 
tokens according to $\pi_{\theta,t}^{\text{unmask}}$. The diffusion 
model then predicts the value of each selected token via 
$p_{\theta,t}(\cdot \mid x_{t-1})$, yielding $x_t^i$ for each 
$i \in \mathcal{U}_t$. The masked set is updated as 
$\mathcal{M}_t = \mathcal{M}_{t-1} \setminus \mathcal{U}_t$, 
and the process repeats for all $T$ steps to produce the 
final sample $x_T$. The complete procedure is summarized 
in Algorithm~\ref{alg:dcollt}.

\begin{algorithm}[th!]
\caption{Training Control Block for Multimodal Diffusion Language Model}
\label{alg:dcollt}
\begin{algorithmic}[1]
\Require Model $\theta$, dataset $\mathcal{D}$, reward function \texttt{reward\_func}
\While{$\theta$ not converged \textbf{and} max epochs not reached}
    \State Sample batch of inputs $c \sim \mathcal{D}$
    \For{$g = 1$ to $G$} 
        \State Initialize $x_0^g$ with $c$ and mask tokens
        \For{$t = 1$ to $T$}
            \State Compute ranking scores $h_{\theta,t}$ for each token
            \State Sample unmask set $\mathcal{U}_t \sim \text{Plackett-Luce}(h_{\theta,t}, K)$
            \State Sample tokens $x_t^{g,i} \sim p_{\theta,t}^i(\cdot \mid x_{t-1}^g)$ for $i \in \mathcal{U}_t$
        \EndFor
        \State Compute reward $r^g \gets \texttt{reward\_func}(c, x_T^g)$
    \EndFor
    \State Compute advantages $A^g$ using group statistics
    \For{$n = 1$ to $T$}
        \State Compute $\pi_{\theta,t}(x_t^g \mid x_{t-1}^g)$
        \State Compute per-step loss $\mathcal{L}_{\theta,t}$ and gradient
    \EndFor
    \State Update $\theta$ using accumulated gradients $\sum_{t=1}^T \nabla_\theta \mathcal{L}_{\theta,t}$
\EndWhile
\end{algorithmic}
\end{algorithm}

\section{Experimental Results}
\label{sec:exp}

In this section, we evaluate our proposed method on compositional image generation and human preference alignment tasks. We first describe the experimental setup, followed by comprehensive benchmark results and ablation studies across both text-to-image generation in Section \ref{sec:exp_t2i} and multimodal understanding in Section \ref{sec:exp_mmu}. Implementation details including optimizer settings, 
hyperparameters, and hardware configuration are provided 
in Appendix~\ref{app:implementation}.

\subsection{Text-to-Image Generation}
\label{sec:exp_t2i}

\paragraph{Compositional Image Generation}
We evaluate our method on the GenEval benchmark~\citep{Ghosh2023GenEvalAO}, a comprehensive suite designed to assess text-to-image (T2I) models under complex compositional prompts involving object counting, spatial relations, and attribute binding. The benchmark comprises six challenging sub-tasks: Single Object, Two Objects, Counting, Colors, Position, and Color Attributes.

Following~\citet{Liu2025FlowGRPOTF}, we generate training prompts using the official GenEval scripts, which apply templates and random combinations to construct a diverse prompt dataset. Evaluation employs the official pipeline for object detection, color identification, and spatial relation inference.

To prevent data leakage, we strictly deduplicate the test set by treating prompts that differ only in object order (e.g., ``a photo of $A$ and $B$'' vs. ``a photo of $B$ and $A$'') as identical variants and removing them from training. Based on the base model's initial task accuracy, we set the prompt sampling ratio as Position : Counting : Attribute Binding : Colors : Two Objects : Single Object = $7 : 5 : 3 : 1 : 1 : 0$.

We employ rule-based rewards for different tasks:
\begin{itemize}[leftmargin=1.5em]
    \item \textbf{Counting:} $r = 1 - \frac{|N_{\text{gen}} - N_{\text{ref}}|}{N_{\text{ref}}}$, providing partial reward when object count is correct;
    \item \textbf{Position/Color:} Full reward when both count and position/color specifications match exactly.
\end{itemize}

Following the same training pipeline as \citep{Liu2025FlowGRPOTF, Li2025MixGRPOUF, Xue2025DanceGRPOUG, He2025TempFlowGRPOWT}, we utilize the mixture reward to learn the control block. We adopt PickScore~\citep{Kirstain2023PickaPicAO} as our reward model to align model outputs with human judgments. PickScore is trained on large-scale human-annotated pairwise comparisons and scores each image-prompt pair based on multiple criteria, including prompt alignment and visual quality.

\paragraph{Results on GenEval}
Table~\ref{tab:geneval} presents our results on the GenEval benchmark. Although confidence-based methods such as Top-K margin perform well on Sudoku puzzles, we find that such strategies do not yield improvement in image generation quality. Even when training with PickScore reward and evaluating on GenEval metrics, our method still achieves superior performance, which demonstrates the generalization ability of the control block. Our approach achieves the best overall performance when trained with in-distribution rewards, representing significant improvements in patial positioning and multi-object composition compared to baseline methods. Notably, we observe substantial gains in the Two Objects (+0.08) and Position (+0.07) tasks, which we attribute to improved generation order control that better handles spatial relationships and multi-object compositions.

\begin{table}[t]
\centering
\caption{Evaluation results on GenEval benchmark. All metrics represent accuracy scores, with higher values indicating better performance. }
\label{tab:geneval}
\resizebox{\linewidth}{!}{
\begin{tabular}{@{}lcccccccc@{}}
\toprule
\multirow{2}{*}{\textbf{Method}} & \multicolumn{7}{c}{\textbf{GenEval Benchmark}} \\
\cmidrule(lr){2-8}
 & \textbf{Single Obj.} & \textbf{Two Obj.} & \textbf{Counting} & \textbf{Colors} & \textbf{Position} & \textbf{Color Attr.} & \textbf{Overall} \\
\midrule
MMaDA-COT Top-K            & 0.93 & 0.47 & 0.31 & 0.81 & 0.16 & 0.26 & 0.49 \\
MMaDA-COT Top-K Margin     & 0.88 & 0.49 & 0.29 & 0.79 & 0.15 & 0.28 & 0.48 \\
MMaDA-Ours (PickScore)     & \textbf{0.95} & 0.45 & \textbf{0.34} & 0.79 & 0.17 & \textbf{0.30} & 0.50 \\
\textbf{MMaDA-Ours (GenEval)} & \textbf{0.95} & \textbf{0.55} & 0.25 & \textbf{0.83} & \textbf{0.23} & 0.26 & \textbf{0.51} \\
\bottomrule
\end{tabular}
}
\end{table}

\subsection{Multimodal Understanding}
\label{sec:exp_mmu}

\paragraph{Benchmark Setup}
We train the control block on the VLAA-thinker dataset \citep{Chen2025SFTOR}, focusing on the \textit{digit} and \textit{multiple-choice question} subsets. The training objective adopts a weighted reward function defined as
\[
\mathcal{R} = 2 \times \mathcal{R}_{\text{correctness}} + 0.4 \times \mathcal{R}_{\text{format}},
\]
where $\mathcal{R}_{\text{correctness}}$ evaluates whether the final answer is correct, and $\mathcal{R}_{\text{format}}$ evaluates whether the response contains a Chain-of-Thought enclosed within the \texttt{<think>} tags.
 We evaluate our method on a comprehensive suite of multimodal benchmarks using the \texttt{VLMEvalKit} framework. This evaluation covers diverse aspects of multimodal reasoning, perception, and language understanding, ensuring robust comparison with existing approaches.

\paragraph{Evaluation Datasets}
We conduct experiments on six widely-adopted benchmarks, each targeting different multimodal capabilities:
\begin{itemize}[leftmargin=1.5em]
    \item \textbf{GQA}~\citep{Hudson2019GQAAN}: Large-scale visual question answering focusing on compositional reasoning over real-world images.
    \item \textbf{MMMU}~\citep{yue2023mmmu}: Multimodal massive multitask benchmark spanning science, mathematics, and humanities domains for knowledge-intensive reasoning evaluation.
    \item \textbf{MMB}~\citep{liu2024mmbench}: Comprehensive multimodal benchmark for language-vision models with diverse visual understanding tasks.
    \item \textbf{SEED}~\citep{ge2024seedxmm}: Multi-granularity comprehension benchmark targeting fine-grained visual-textual alignment.
    \item \textbf{MathVista\_MINI}~\citep{lu2024mathvista}: Challenging multimodal mathematical reasoning dataset requiring symbolic and visual integration.
    \item \textbf{COCO\_VAL}~\citep{Lin2014MicrosoftCC}: MS-COCO validation split for evaluating image-text understanding and generation consistency.
\end{itemize}

\paragraph{Multimodal Understanding Results}
Table~\ref{tab:multimodal} summarizes quantitative results across all evaluated datasets. Similarly, confidence-based methods such as Top-K margin do not achieve consistent improvement over the baseline method. Our approach achieves the best results on four of six benchmarks. Consistent improvements are observed on MathVista\_MINI (+1.3) and COCO\_VAL (+0.7), both of which involve open-ended questions and complex reasoning processes compared to multiple-choice benchmarks. These results validate that optimizing generation order effectively enhances multimodal understanding and reasoning capabilities.

\begin{table}[t]
\centering
\caption{Evaluation results on multimodal understanding benchmarks.}
\label{tab:multimodal}
\resizebox{\linewidth}{!}{
\begin{tabular}{@{}lcccccc@{}}
\toprule
\textbf{Method} & \textbf{GQA} & \textbf{MMMU} & \textbf{MMB} & \textbf{SEED} & \textbf{MathVista\_MINI} & \textbf{COCO\_VAL} \\
\midrule
MMaDA-COT                & \textbf{48.58} & 0.27 & 0.24 & 0.52 & 27.7 & 27.6 \\
MMaDA-COT Top-K Margin   & 47.55 & 0.24 & \textbf{0.37} & 0.54 & 28.7 & 26.3 \\
\textbf{MMaDA-Ours}      & 47.61 & \textbf{0.31} & 0.24 & \textbf{0.57} & \textbf{29.0} & \textbf{28.3} \\
\bottomrule
\end{tabular}
}
\end{table}

\subsection{Ablation Studies and Analysis}
\label{sec:ablation}

\begin{figure}[h!]
\centering
  \includegraphics[width=0.9\textwidth]{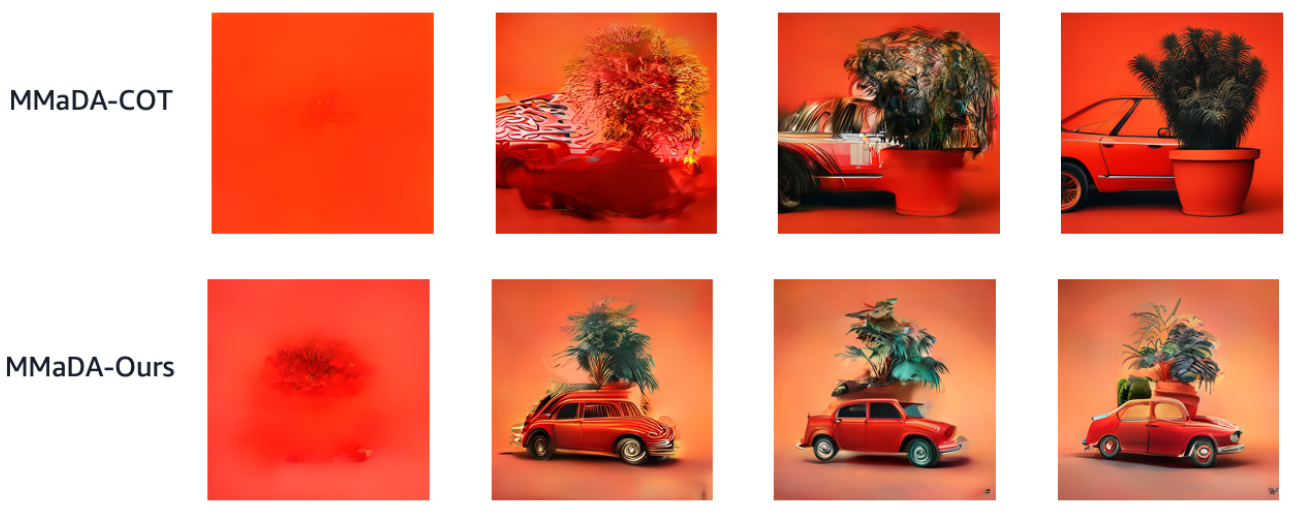}
  \caption{Trajectory visualization on text-to-image generation.}
  \label{fig:trajectory_t2i}
\end{figure}

\begin{figure}[h!]
\centering
  \includegraphics[width=0.9\textwidth]{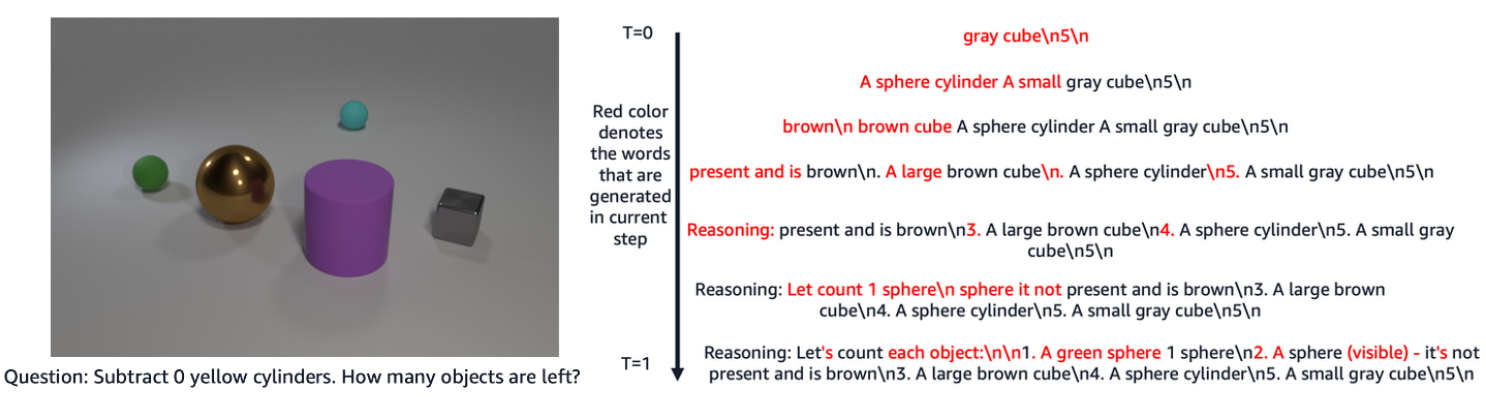}
  \caption{Trajectory visualization on multimodal understanding.}
  \label{fig:trajectory_mmu}
\end{figure}

\paragraph{Generation Order Optimization Analysis}
To better understand the effect of our method on the generation process, we visualize the generation trajectories for both text-to-image generation and multimodal understanding tasks in Figure~\ref{fig:trajectory_t2i} and Figure~\ref{fig:trajectory_mmu}. For text-to-image generation, our approach improves intermediate results across the trajectory, leading to higher-quality final outputs. For multimodal understanding, while the baseline model—despite being allowed up to 256 output tokens—often degenerates to a single-token response (e.g., “5” illustrated in Figure \ref{fig:trajectory_mmu}), our control block enables more detailed reasoning. The model is able to correctly solve counting problems and explicitly name objects to support its reasoning. These results demonstrate that optimizing the generation order not only enhances the model’s ability to capture fine-grained spatial relationships in images but also strengthens its performance on multimodal reasoning and comprehension tasks.

\begin{figure}[h]
\centering
\begin{minipage}[t]{0.48\textwidth}
    \centering
    \includegraphics[width=\textwidth]{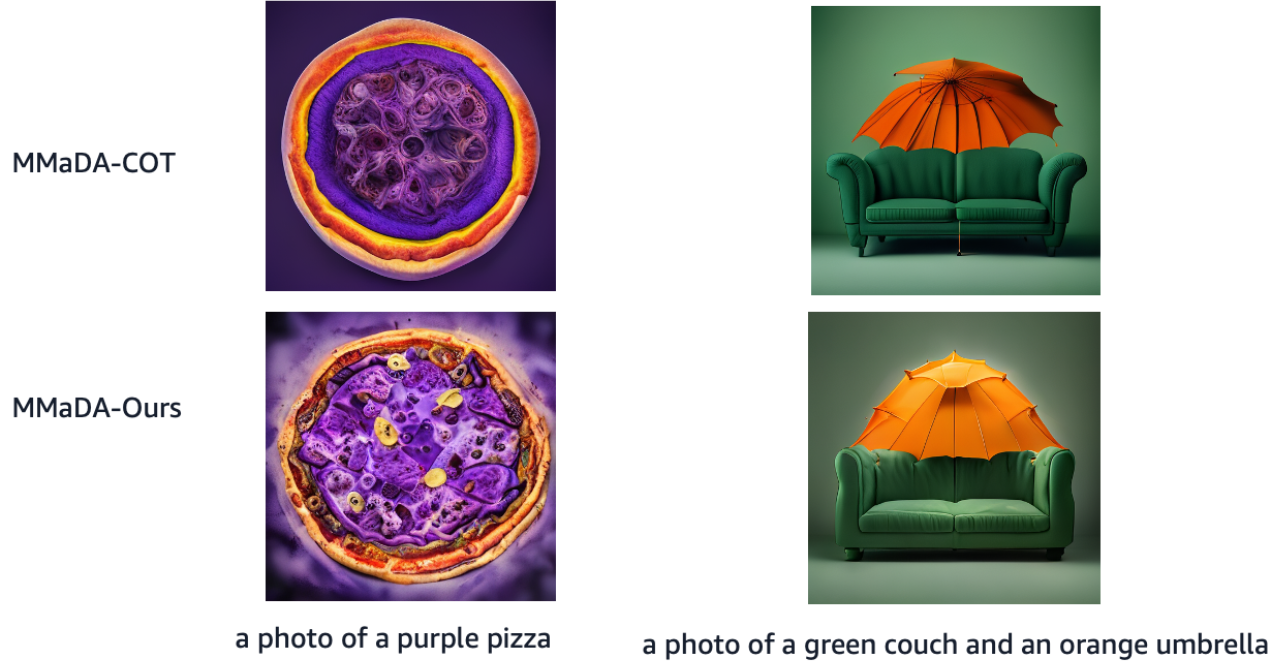}
    \subcaption{Text-to-Image Generation Results}
    \label{fig:qual_t2i}
\end{minipage}
\hfill
\begin{minipage}[t]{0.48\textwidth}
    \centering
    \includegraphics[width=\textwidth]{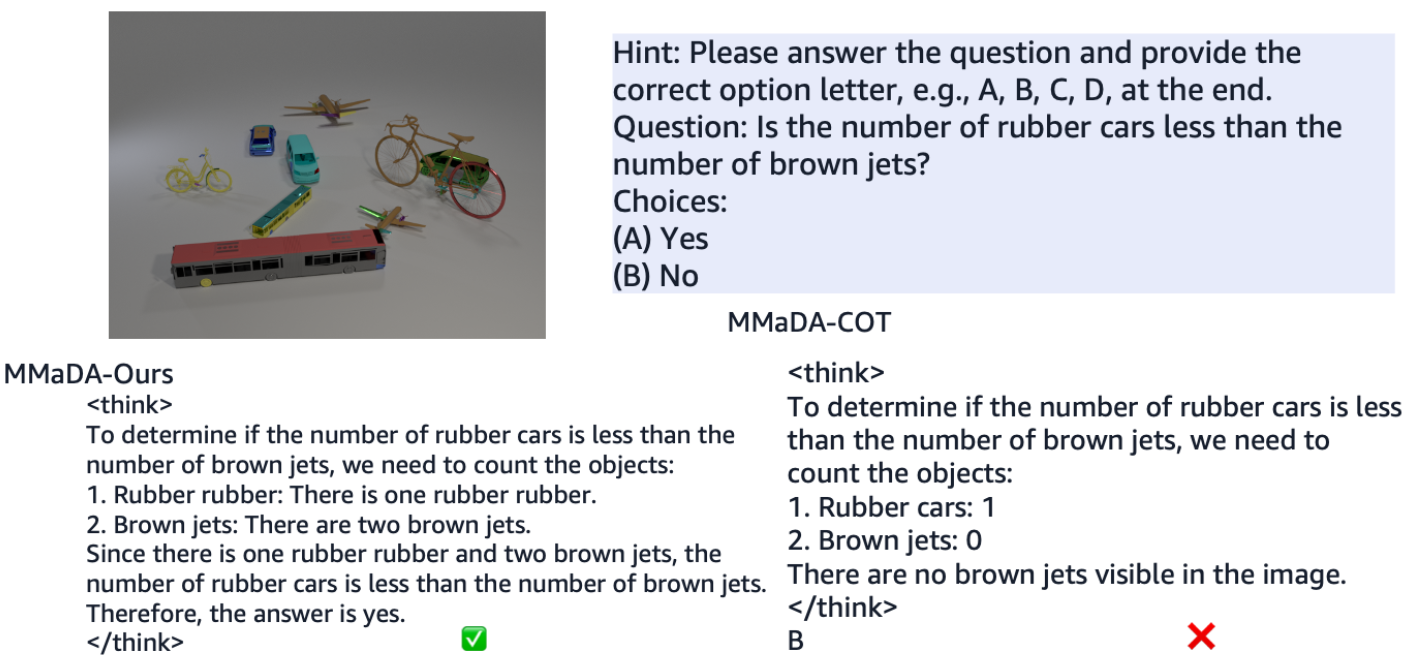}
    \subcaption{Multimodal Understanding Results }
    \label{fig:qual_mmu}
\end{minipage}
\caption{Qualitative comparison of generated samples. Our method demonstrates superior understanding and generation quality across both text-to-image synthesis and multimodal reasoning tasks.}
\label{fig:qualitative_comparison}
\end{figure}

\paragraph{Qualitative Comparison}
Figure~\ref{fig:qualitative_comparison} presents qualitative comparisons of generated samples for both text-to-image generation and multimodal understanding tasks. Our method produces more coherent and contextually appropriate outputs, particularly in scenarios requiring complex compositional reasoning.

\section{Related work}
\paragraph{Masked diffusion models}
Masked diffusion models (discrete diffusion model with absorbing state) were first proposed by extending the transition probability from a Gaussian distribution used in the continuous diffusion model to a multinormal distribution \citep{SohlDickstein2015DeepUL, Austin2021StructuredDD, Hoogeboom2021ArgmaxFA}. Then, continuous Time framework was established by \cite{Campbell2022ACT, Sun2022ScorebasedCD}. A line of works \citep{Lou2023DiscreteDM, Arriola2025BlockDI, Shi2024SimplifiedAG, Ou2024YourAD, Zheng2024MaskedDM, Sahoo2024SimpleAE, Rutte2025GeneralizedID} proposed simplified objective to train masked diffusion model. Discrete flow matching was proposed in \cite{Gat2024DiscreteFM, Campbell2024GenerativeFO}. In \cite{Nie2025LargeLD}, the authors scale up the masked diffusion model to 8B, and Yang et al. \citep{Yang2025MMaDAML, Xin2025LuminaDiMOOAO, Wang2025FUDOKIDF} generalize the masked diffusion model and discrete flow matching on multimodal tasks. 

\paragraph{Control over generation order}
Recent work has increasingly recognized the importance of studying the generation order for masked diffusion model \citep{Kim2025TrainFT, Huang2025ReinforcingTD}. Instead of randomly selecting a set of positions to generate, Zheng et al. \citep{Zheng2023ARD} propose Top K strategy, i.e., the generation order is selected by the maximum probability of predicted token in each position. In \cite{Kim2025TrainFT}, the authors propose top K margin strategy, i.e.,  the absolute difference between the two most probable values at each position. They show that by leveraging the information for the logits, they can dramatically improve the performance of MDM on math puzzles like Sudoku. 




\section{Conclusion}\label{sec:conclusion}
We have investigated the problem of generation order in multimodal masked diffusion models for text-to-image generation. While prior logits-based strategies such as Top-$K$ margin are effective in language reasoning tasks, we showed that they fail to improve image generation quality. To overcome this, we proposed a learnable control block optimized with Group Relative Policy Optimization (GRPO), enabling adaptive and task-aware order selection. Experiments on the GenEval benchmark and VLMEvalKit benchmarks demonstrate improvements over existing baselines, confirming the importance of learning-based order policies in multimodal diffusion models.

\bibliographystyle{plainnat}
\bibliography{ref}

\appendix

\section{Implementation Details}
\label{app:implementation}

We build upon \texttt{MMaDA-8B-MixCoT}~\citep{Yang2025MMaDAML} 
as our base model and optimize the control block via a 
GRPO-style reinforcement learning objective. For all experiments, 
we use the AdamW optimizer with a constant learning rate of 
$5 \times 10^{-6}$, $\beta_1 = 0.9$, $\beta_2 = 0.999$, 
$\epsilon = 10^{-8}$, and no weight decay. All models are 
trained for $5{,}000$ steps with a per-GPU batch size of $1$ 
and $4$ gradient accumulation steps. 
The clipping coefficient is set to $\epsilon_{\text{clip}} = 0.2$ 
and the KL penalty coefficient to $\beta = 0$.

For text-to-image generation (GenEval), experiments are 
conducted on $24$ NVIDIA A100 GPUs. We sample $G = 4$ 
completions per prompt with $R = 6$ repeat sampling rounds 
per training step (i.e., $24$ total samples per prompt), 
using $16$ denoising steps and a generation length of 
$1{,}024$ tokens. Training is conducted in \texttt{bfloat16} 
precision.

For multimodal understanding (VLMEvalKit), experiments are 
conducted on $8$ NVIDIA A100 GPUs. We sample $G = 4$ 
completions per prompt with $R = 3$ repeat sampling rounds 
(i.e., $12$ total samples per prompt), using $64$ denoising 
steps and a generation length of $256$ tokens, trained in 
\texttt{float16} precision.
\end{document}